\pdfoutput=1
\documentclass[11pt, letterpaper, logo, onecolumn, copyright]{template_from_google}

\usepackage[T1]{fontenc}
\usepackage{amsmath}
\usepackage{amsfonts}
\usepackage{amssymb}
\usepackage{amsthm}
\usepackage{nicefrac}
\usepackage{booktabs}
\usepackage{enumitem}
\usepackage{graphicx}
\usepackage{microtype}
\usepackage[dvipsnames]{xcolor}
\usepackage{url}
\usepackage[authoryear,sort&compress,round]{natbib}
\usepackage{newunicodechar}
\usepackage{array}
\newunicodechar{−}{-}
\newunicodechar{–}{--}
\newunicodechar{—}{---}
\newunicodechar{×}{\ensuremath{\times}}
\newunicodechar{∼}{\ensuremath{\sim}}
\newunicodechar{≈}{\ensuremath{\approx}}
\newunicodechar{≤}{\ensuremath{\leq}}
\newunicodechar{≥}{\ensuremath{\geq}}
\newunicodechar{≠}{\ensuremath{\neq}}
\newunicodechar{→}{\ensuremath{\to}}
\newunicodechar{←}{\ensuremath{\leftarrow}}
\newunicodechar{α}{\ensuremath{\alpha}}
\newunicodechar{β}{\ensuremath{\beta}}
\newunicodechar{γ}{\ensuremath{\gamma}}

\hypersetup{
  pdffitwindow=true,
  pdfstartview={FitH},
  pdfnewwindow=true,
  colorlinks=true,
  linktocpage=true,
  linkcolor=MidnightBlue,
  urlcolor=MidnightBlue,
  citecolor=MidnightBlue
}

\newcommand{\sem}{\textsuperscript{S}}
\newcommand{\clus}{\textsuperscript{C}}
\newcommand{\diffm}{\textsuperscript{D}}
\newcommand{\drop}{\textsuperscript{T}}

\newtheorem{proposition}{Proposition}
\newtheorem*{remark}{Remark}

\title{Taming the Entropy Cliff: Variable Codebook Size Quantization for Autoregressive Visual Generation}

\author[1]{Bowen Zheng}
\author[2]{Weijian Luo}
\author[2]{Guang Yang}
\author[2]{Colin Zhang}
\author[1]{Tianyang Hu}

\affil[1]{The Chinese University of Hong Kong, Shenzhen}
\affil[2]{hi-Lab, Xiaohongshu Inc}

\reportnumber{}
\renewcommand{\today}{}
\newcommand{\shorttitle}{Variable Codebook Size Quantization for Autoregressive Visual Generation}
\begin{abstract}
Most discrete visual tokenizers rely on a default design: every position in the sequence shares the same codebook. Researchers try to scale the codebook size $K$ to get better reconstruction performance. Such a constant-codebook design can consume the finite-data entropy budget very quickly. We observe that the per-position empirical conditional entropy decays so quickly along the sequence that, after a few positions, the conditional distribution becomes essentially deterministic. On ImageNet with $K=16384$, this happens after only about 2 out of 256 positions, when most prefixes already identify individual images in the dataset. We call this phenomenon the \textbf{Entropy Cliff}. Under full conditional utilization, larger codebooks exhaust the finite-data entropy budget in fewer positions.
Interestingly, the empirical entropy decreases more gradually in our language comparison, where the effective entropy per position stays below the codebook capacity. To address this, we propose \textbf{Variable Codebook Size Quantization (VCQ)}, where the codebook size $K_t$ grows monotonically along the sequence from $K_{\min}=2$ to $K_{\max}$, leaving the loss function, parameter count, and AR training procedure unchanged. With a vanilla autoregressive Transformer and standard next-token prediction, VCQ-Base achieves \textbf{gFID w/o CFG 14.80} on ImageNet $256\times256$, compared with 27.98 and 18.16 for constant codebooks of size 16384 and 8192, respectively. Scaled up, it reaches \textbf{gFID 1.71} with 684M autoregressive parameters, without any extra training techniques such as semantic regularization or causal alignment. The extreme information bottleneck at $K_{\min}=2$ encourages semantic concentration in early encoder features: a linear probe on features from only the first 10 positions reaches \textbf{43.8\% top-1 accuracy} on ImageNet, compared to 27.1\% for uniform codebooks. Ultimately, these results show that what matters is not only the total capacity of the codebook, but also how that capacity is distributed and organized.
 \end{abstract}

\begin{document}
\maketitle

\section{Introduction}
Discrete tokenizers are the foundation of autoregressive visual generation. From VQ-VAE~\citep{vqvae} and VQ-GAN~\citep{vqgan} to more recent works such as LlamaGen~\citep{llamagen}, GigaTok~\citep{gigatoken}, a persistent trend has been to increase the codebook size $K$ for better reconstruction quality; a size of 16k or more has become standard. These works differ in training objectives, network architectures, and inference strategies, yet they share one design choice with striking unanimity: every position in the sequence can potentially use all tokens in the same codebook size. We refer to this default codebook setting as the \textbf{\emph{constant codebook size}} quantization or uniform codebooks.

\begin{figure}[!t]
\centering
\includegraphics[width=\textwidth]{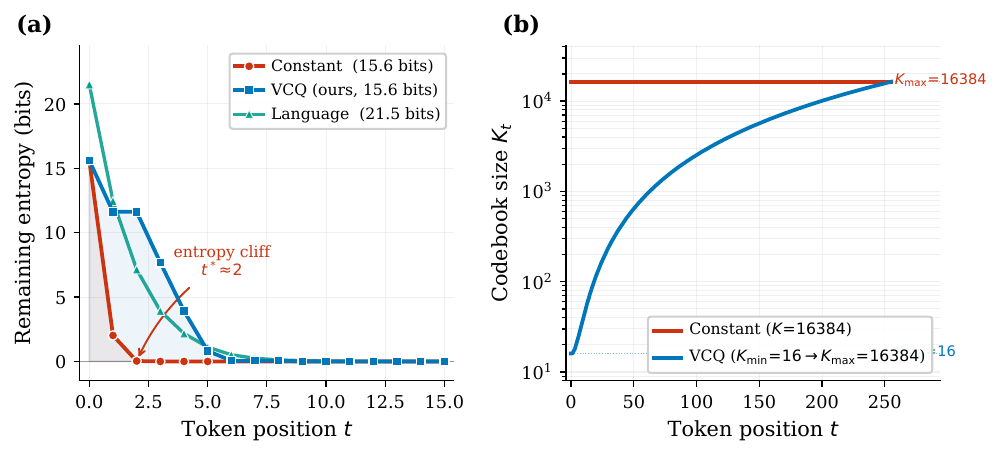}
\caption{\textbf{Entropy cliff and codebook schedule.}
(a)~Empirical remaining budget estimated from token-prefix frequencies. The image curves are measured on ImageNet. Constant ($K\!=\!16384$) approaches zero after about 2 positions, while VCQ spreads the decrease across more positions. The LAMBADA comparison decreases more gradually. For clarity, this plot uses a Power schedule with $K_{\min}\!=\!16$.
(b)~Codebook size schedule: Constant is fixed at $K\!=\!16384$; VCQ increases gradually from $K_{\min}$ to $K_{\max}$.}
\label{fig:entropy}
\end{figure}

However, the uniform codebook has significant flaws (Fig.~\ref{fig:entropy}(a)). We encode ImageNet images into 256 tokens with a uniform codebook of $K\!=\!16384$ (the codebook from LlamaGen~\citep{llamagen}) and calculate empirical conditional entropy from prefix frequencies. At $t\!=\!0$, the conditional entropy is about 14 bits. By $t\!=\!2$, it falls below 1 bit and approaches zero thereafter. Most observed prefixes then identify individual images in the dataset, each with a fixed empirical continuation. We call this phenomenon the \textbf{\emph{Entropy Cliff}}. On the full ImageNet training set, 96.6\% of images already have unique two-token prefixes (Appendix~\ref{app:prefix_counts}).

Regularization such as weight decay acts on the AR model, while codebook size controls the tokenized representation. A naive remedy is to globally reduce the codebook size, which limits early-position capacity and can delay the empirical cliff.
However, uniformly shrinking the codebook also restricts later positions, where richer local detail is required, which can reduce reconstruction fidelity. This motivates controlling how capacity is distributed across the autoregressive sequence. Early positions primarily determine high-level semantic structure and are particularly sensitive to excessive discrete freedom, while later positions benefit from larger codebooks to model fine-grained details. 

This observation motivates \textbf{Variable Codebook Size Quantization (VCQ)}, which assigns smaller codebooks to early positions to stabilize global autoregressive generation, and progressively increases the codebook size at later positions to preserve reconstruction quality and local detail. Specifically, $K_t$ increases monotonically from $K_{\min}\!=\!2$ to $K_{\max}$ (Eq.~\ref{eq:schedule}). At early positions, smaller codebooks impose a stronger information bottleneck that delays the onset of the entropy cliff; at later positions, larger codebooks provide sufficient capacity to preserve reconstruction quality.

\begin{figure}[!t]
\centering
\includegraphics[width=\textwidth]{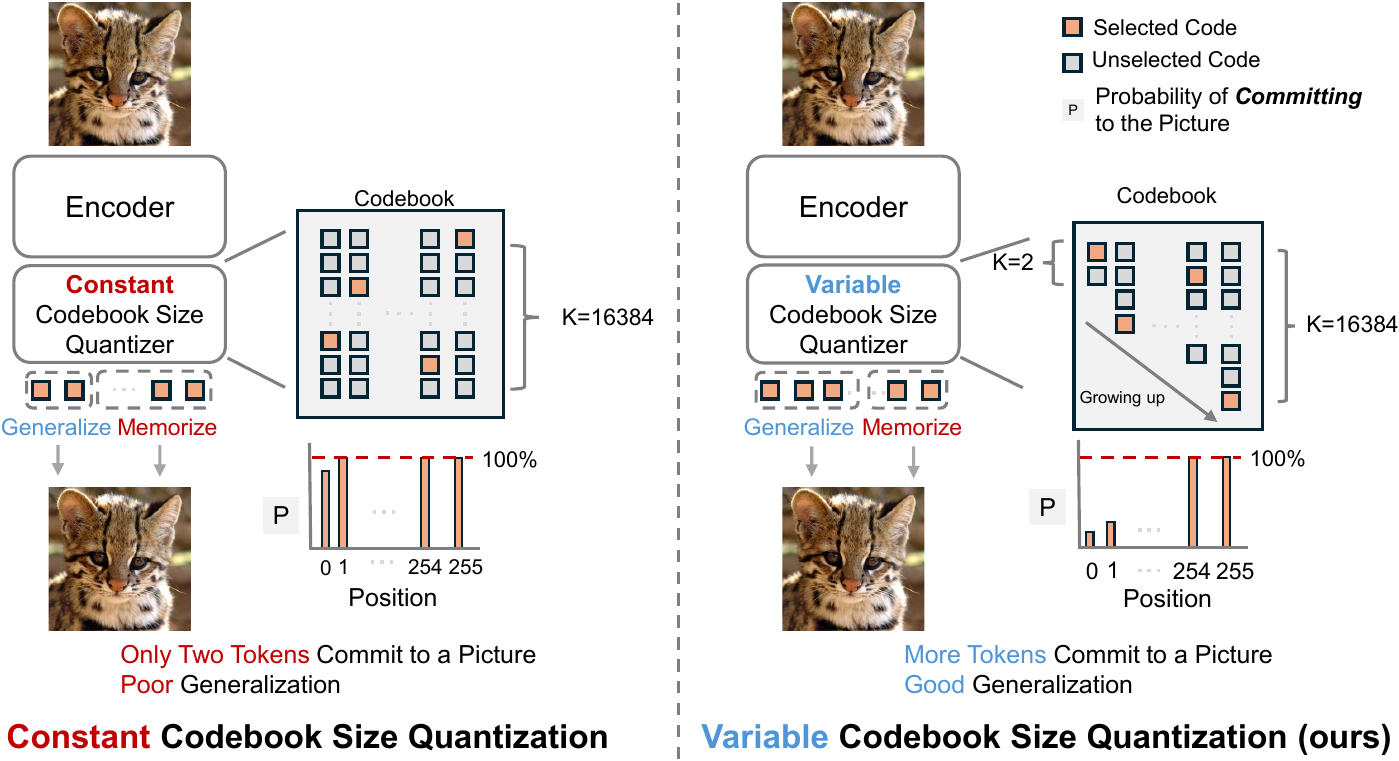}
\caption{\textbf{VCQ architecture.} \textbf{Left}: Constant Codebook Size Quantization, where every position $t$ shares the same codebook of size $K$, with a full-utilization capacity floor of $t^{*}_{\mathrm{floor}}\!=\!\lceil \log_2 N / \log_2 K \rceil$ positions. \textbf{Right}: our Variable Codebook Size Quantization (VCQ), where the codebook size $K_t$ grows monotonically from $K_{\min}\!=\!2$ to $K_{\max}$ along the sequence. Early positions use small codebook size to impose a strong information bottleneck and delay the entropy cliff, while later positions use larger codebook size to preserve reconstruction fidelity. We implement VCQ with a single shared codebook of $K_{\max}$ entries, where position $t$ selects the nearest neighbor from only the first $K_t$ entries—introducing no extra parameters and leaving the loss function and AR training procedure unchanged.}
\label{fig:overview}
\end{figure}

We observe advantages of VCQ along three axes:
\begin{enumerate}[leftmargin=*,nosep]
\item \textbf{Delayed entropy cliff.} VCQ spreads the empirical entropy across more positions by limiting early-position capacity. The finite-data analysis in \S\ref{sec:cliff} explains how the codebook schedule changes the capacity floor, while the actual entropy curve also depends on conditional utilization.
\item \textbf{Improved generation quality.} Compared with the constant $K=16384$ baseline, VCQ-Base reduces gFID w/o CFG from 27.98 to 14.80 ($-47\%$), and, with CodebookSize-Aware CFG, gFID from 6.43 to 4.79 ($-26\%$). Generation quality improves with scale: VCQ-Large achieves gFID=1.71 with a 684M-parameter AR model, without semantic alignment or causal regularization (\S\ref{sec:generation}).
\item \textbf{Semantic concentration.} $K_{\min}\!=\!2$ encourages global semantic information to concentrate in early encoder features. Using features from only the first 10 positions (3.9\% of the sequence), linear probing top-1 accuracy on ImageNet classification reaches 43.8\%, compared to 27.1\% for constant-codebook (\S\ref{sec:semantic}).
\end{enumerate}

\section{The entropy cliff of constant codebook size quantization}
\label{sec:cliff}

\textbf{Notation.} We consider the empirical distribution of $N$ equally weighted tokenized samples with sequence length $L$. Positions are indexed from $0$. Position $t$ has at most $K_t$ possible codes, and $H_{\mathrm{emp}}(x_t\mid x_{<t})$ denotes its empirical conditional entropy.

\subsection{A finite-data entropy budget}
Define the remaining budget after a length-$t$ prefix as
\begin{equation}
 R(t)=\log_2 N-\sum_{i<t}H_{\mathrm{emp}}(x_i\mid x_{<i}).
 \label{eq:remaining}
\end{equation}
\Needspace{7\baselineskip}
\begin{proposition}[Finite-data entropy budget]
\label{prop:cliff}
Let $I(t)=\sum_{i<t}\log_2 K_i$. Then
\begin{equation}
 \max\bigl(0,\log_2 N-I(t)\bigr)\leq R(t)\leq\log_2 N,
 \label{eq:cuminfo}
\end{equation}
and the entire remaining token sequence has empirical conditional entropy at most $R(t)$. If all length-$t$ prefixes are distinct, $R(t)=0$ and all subsequent empirical continuations are deterministic.
\end{proposition}
The proof and full derivation are in Appendix~\ref{app:entropy_theory}.

For a constant codebook, the capacity floor reaches zero at
\begin{equation}
 t^*_{\mathrm{floor}}=\left\lceil\frac{\log_2 N}{\log_2 K}\right\rceil.
 \label{eq:tstar}
\end{equation}
To describe the actual decay, let $A_t=\min(\log_2K_t,R(t))$ and define conditional utilization as $\rho_t=H_{\mathrm{emp}}(x_t\mid x_{<t})/A_t\in[0,1]$ for $A_t>0$, with $\rho_t=0$ otherwise. Then $R(t+1)=R(t)-\rho_t A_t$. Full conditional utilization ($\rho_t=1$ whenever $A_t>0$) makes the budget follow the capacity floor. Lower utilization can delay the measured cliff.

\paragraph{Information redistribution by VCQ.}
VCQ keeps $K_t$ small at early positions and increases it toward $K_{\max}$. This slows the growth of $I(t)$ and raises the early-position floor on $R(t)$. Under full conditional utilization, the floor reaches zero at the first $t$ for which $I(t)\geq\log_2N$. VCQ therefore changes the data-side capacity constraint directly, without changing the AR training objective.

\paragraph{Empirical measurement.}
\label{sec:entropy_verify}
For each observed prefix $p=x_{<t}$, we count its occurrences $n(p)$ and the occurrences $n(p,c)$ followed by code $c$. We compute
\begin{equation}
 H_{\mathrm{emp}}(x_t\mid x_{<t})=-\sum_{p,c}\frac{n(p,c)}{N}\log_2\frac{n(p,c)}{n(p)},
 \label{eq:entropy_estimator}
\end{equation}
where terms with zero count are omitted. Figure~\ref{fig:entropy}(a) shows the remaining budget derived from these empirical counts. Its image curves are measured on ImageNet. The constant-codebook curve approaches the capacity floor, while VCQ delays the decrease.

\paragraph{Training-set evidence.}
On the full ImageNet training set of 1.28M images, 96.6\% have unique two-token prefixes with $K=16384$. After three tokens, the empirical next-token entropy falls below 0.002 bits. These counts directly support the entropy cliff in the training data: most prefixes have only one observed continuation (Appendix~\ref{app:prefix_counts}, Table~\ref{tab:prefix_counts}).

\begin{remark}[Language comparison]
We also measure 3M sliding-window samples of length 256 from LAMBADA~\citep{lambada}, using the Llama~\citep{llama} tokenizer with a vocabulary of approximately 32K. At the first position, $\rho_0$ is 97.1\% for the constant-codebook image tokenizer and 60.5\% for language. Language consumes a smaller fraction of the available budget initially, and its measured budget decreases more gradually.
\end{remark}

\subsection{The data-side problem and the role of regularization}
\label{sec:cliff_gen}
Once an observed prefix uniquely identifies a sample, its empirical continuation is fixed. When this happens at most positions, repeated-prefix evidence about conditional variability is largely absent from the finite dataset. This is the data-side problem identified by the entropy cliff.

Regularization and shared parameters allow the model to learn a smoother solution across different prefixes. Consistent with this, the baseline's predictive entropy on ImageNet averages 13.5 bits over positions 3--16 and never falls below 9.1 bits. Regularization smooths the model distribution, while the empirical continuation counts describe the tokenized data.

VCQ addresses the data side by restricting early codebooks and spreading the empirical information across more positions. Under the same AR training configuration and weight decay of 0.03, VCQ-Cosine improves gFID w/o CFG from 27.98 to 14.80 and first-10-position probing accuracy from 27.1\% to 43.8\%. These results support changing the tokenizer's information allocation while retaining the regularization used by the baseline.

\section{Variable codebook quantization}
\label{sec:vcq}
Fig.~\ref{fig:overview} illustrates the difference between Constant Codebook Size Quantization and our Variable Codebook Size Quantization. Our method is simple and intuitive: the only change is to make the codebook size vary by position rather than remain fixed. We elaborate on two design choices below.
\paragraph{Schedule family}
$K_t$ increases monotonically from $K_{\min}$ to $K_{\max}$:
\begin{equation}
  K_t = K_{\min} + (K_{\max} - K_{\min}) \cdot f\!\left(\frac{t}{L{-}1}\right), \qquad f(0)=0,\; f(1)=1
  \label{eq:schedule}
\end{equation}
$f$ controls the convexity of the growth curve (Fig.~\ref{fig:entropy}(b)):
\begin{itemize}[leftmargin=*,nosep]
\item \textbf{Linear}: $f(\tau) = \tau$. Uniform growth, no hyperparameters.
\item \textbf{Cosine}: $f(\tau) = 1 - \cos(\pi \tau / 2)$. Slow at the beginning, fast toward the end. \hfill \textit{(default)}
\item \textbf{Power}: $f(\tau) = \tau^\alpha$. $\alpha$ controls the degree of suppression at early positions.
\end{itemize}
Table~\ref{app:schedule} summarizes the concrete schedule configurations used in our VCQ experiments, including the minimum and maximum codebook sizes, the resulting average codebook size $\bar{K}$, and the nominal bits per pixel (BPP). Slower growth at early positions places a larger share of capacity at later positions, at the cost of a smaller average codebook size $\bar{K}$ and higher rFID. The uniform codebook is the special case $K_{\min}=K_{\max}$. We default to $K_{\min}\!=\!2$, $K_{\max}\!=\!16384$, and the Cosine schedule.

\begin{table}[h]
\caption{Schedule parameters. $\bar{K}$ is the average codebook size, and BPP the nominal bits per pixel.}\label{app:schedule}
\begin{center}
\small
\begin{tabular}{lcccc}
\toprule
Schedule & $K_{\min}$ & $K_{\max}$ & $\bar{K}$ & BPP \\
\midrule
Constant ($K$=16384) & 16384 & 16384 & 16384 & 0.055 \\
Constant ($K$=8192)  & 8192  & 8192  & 8192  & 0.051 \\
Linear               & 2     & 16384 & 8193  & 0.049 \\
Cosine               & 2     & 16384 & 5964  & 0.044 \\
Power ($\alpha$=2.5) & 2     & 16384 & 4696  & 0.041 \\
Cosine-L             & 2     & 11264 & 4100  & 0.042 \\
\bottomrule
\end{tabular}
\end{center}
\end{table}

\paragraph{Shared codebook}
 We implement VCQ with a shared codebook: a single global codebook of $K_{\max}$ entries is maintained, and position $t$ selects the nearest neighbor from only the first $K_t$ entries. No extra parameters are introduced and the gradient flow is unchanged. The tokenizer is trained with standard VQ losses~\citep{vqgan} (L1+LPIPS~\citep{lpips}+PatchGAN~\citep{pix2pix}) and the AR model with standard next-token prediction. VCQ changes only the size of the quantization candidate set at each position. Our GPU implementation computes distances to all $K_{\max}$ codes before masking invalid entries, retaining the full-codebook search cost.

\paragraph{CodebookSize-Aware CFG}
The non-uniform codebook of VCQ requires a corresponding adjustment to CFG: at early positions where the codebook is small and the distribution is concentrated, standard CFG over-conditions. CodebookSize-Aware CFG ties the guidance scale to the excess information capacity at position $t$:
\begin{equation}
  s_t = \bar{s}_t \cdot \frac{\log_2 K_t - \log_2 K_{\min}}{\log_2 K_{\max} - \log_2 K_{\min}}
  \label{eq:bitcfg}
\end{equation}
where $\bar{s}_t$ is the base CFG scale at position $t$, following the cosine-power schedule in Appendix~\ref{app:inference}. When $K_t = K_{\min}$, $s_t = 0$ (no CFG); when $K_t = K_{\max}$, $s_t = \bar{s}_t$ (standard CFG). The output logits at position $t$ are $(1 + s_t)\,\text{logits}_{\text{cond}} - s_t\,\text{logits}_{\text{uncond}}$. For constant codebooks, we use cosine CFG with the settings in Appendix~\ref{app:inference}. Appendix~\ref{app:posaware} shows that lowering its initial scale from 1 to 0 brings no clear improvement.

\section{Experiments}
\label{sec:exp}

\subsection{Setup}
We conduct our main experiment on ImageNet 256$\times$256~\citep{ilsvrc}. Our baseline tokenizer is a classical 1D tokenizer~\citep{titok} with sequence length=256. We train the tokenizer with standard L1, LPIPS and PatchGAN loss. The tokenizer is trained without class labels; class conditioning is introduced only during AR training. The AR model is then trained on the frozen tokenizer with a next-token prediction task. We use the same standard training setup across codebook configurations. All experiments use AdamW~\citep{adamw}. Reconstruction quality is reported with rFID and PSNR; generation quality is reported with gFID~\citep{fid}, Inception Score~\citep{is}, and the Precision/Recall metric~\citep{precrecall}. rFID is reported on the ImageNet validation set, and gFID follows the ADM evaluation protocol. Full configurations are listed in Appendix~\ref{app:training}; throughout the paper, configurations are named as ``Tok-AR'' (e.g., VCQ B-L = Base Tokenizer + Large AR).

\textbf{VCQ configuration.} $K_{\min}\!=\!2$, default Cosine schedule. Base Tok: $K_{\max}\!=\!16384$. Large Tok: $K_{\max}\!=\!11264$.

\begin{table}[h]
  \caption{\textbf{Model configurations.} Tokenizer and AR model variants used in experiments.}
  \label{tab:config}
  \begin{center}
\small
\begin{tabular}{@{}llc@{}}
\toprule
\multicolumn{3}{l}{\textit{Tokenizer}} \\
\midrule
Base  & ViT-B Enc + ViT-B Dec, $K_{\max}\!=\!16384$ & 172M \\
Large & ViT-B Enc + ViT-L Dec, $K_{\max}\!=\!11264$ & 496M \\
\midrule
\multicolumn{3}{l}{\textit{AR Model}} \\
\midrule
B  & & 115M \\
L  & & 343M \\
XL & & 684M \\
\bottomrule
\end{tabular}
\end{center}
\end{table}

\begin{table}[!ht]
  \caption{\textbf{Baseline comparison}. All configurations share exactly the same training setup; the only difference is the codebook structure. gFID uses each method's best inference hyperparameters.}
  \label{tab:internal}
  \begin{center}
\small
\begin{tabular}{@{}lcccccc@{}}
    \toprule
    Schedule & $\bar{K}$ & BPP & PSNR $\uparrow$ & rFID $\downarrow$ & gFID $\downarrow$ & gFID\textsubscript{noCFG} $\downarrow$ \\
    \midrule
    Constant ($K$=16384) & 16384 & 0.055 & \textbf{20.67} & \textbf{2.26} & 6.43          & 27.98 \\
    Constant ($K$=8192)  & 8192  & 0.051 & 20.48          & 2.36          & 5.87          & 18.16 \\
    VCQ-Linear           & 8193  & 0.049 & 20.35          & 2.42          & 5.32          & \textbf{14.70} \\
    VCQ-Cosine           & 5964  & 0.044 & 20.02          & 2.77          & \textbf{4.79} & 14.80 \\
    VCQ-Power ($\alpha$=2.5) & 4696 & 0.041 & 19.82       & 2.89          & 5.14          & 16.19 \\
    \bottomrule
  \end{tabular}
\end{center}
\end{table}

\begin{table}[p]
  \caption{\textbf{Reconstruction and generation comparison} (ImageNet $256^2$ class-conditional).
  Naming format ``Tok-AR'' (B=Base, L=Large, XL=XLarge).
  rFID/PSNR are tokenizer reconstruction properties and are identical across rows that share the same tokenizer.
  ``Diff.''=diffusion, ``M+D''=mask+diffusion, ``Mask.''=masked transformer,
  ``HAR''=hierarchical AR, ``PAR''=parallel AR.
  $^\mathrm{S}$Semantic/VFM: semantic regularization, semantic alignment, or frozen/pre-trained vision foundation model;
  $^\mathrm{C}$Clustering: periodic token/codebook clustering;
  $^\mathrm{D}$Diff./FM: diffusion or flow-matching generator, decoder, token-distribution model, or loss;
  $^\mathrm{T}$Dropout: tail dropout.}
  \label{tab:gen}
  \begin{center}
\fontsize{9}{10.5}\selectfont
\setlength{\tabcolsep}{2pt}
\begin{tabular}{@{}>{\raggedright\arraybackslash}p{175pt}>{\raggedright\arraybackslash}p{65pt}ccccccc@{}}
  \toprule
  Method & \shortstack[l]{Gen \&\\Param}
  & rFID $\downarrow$ & PSNR $\uparrow$
  & gFID $\downarrow$ & \shortstack{gFID\\w/o CFG $\downarrow$}
  & IS $\uparrow$ & Prec $\uparrow$ & Rec $\uparrow$ \\
  \midrule

  \multicolumn{9}{l}{\textit{Continuous tokens (diffusion / MAR)}} \\
  LDM-4-G~\citep{ldm}\diffm
    & Diff.\ 400M   & 0.27 & --- & 3.60 & 10.56 & 247.7 & 0.87 & 0.48 \\
  DiT-XL/2~\citep{dit}\diffm
    & Diff.\ 675M   & 0.62 & --- & 2.27 & 9.62  & 278.2 & 0.83 & 0.57 \\
  SiT-XL~\citep{sit}\diffm
    & Diff.\ 675M   & ---  & --- & 2.06 & 8.61  & 270.3 & 0.82 & 0.59 \\
  REPA~(SiT-XL)~\citep{repa}\sem\diffm
    & Diff.\ 675M   & ---  & --- & 1.42 & 5.90  & 305.7 & 0.80 & 0.65 \\
  LightningDiT~\citep{lightningdit}\sem\diffm
    & Diff.\ 675M   & 0.28 & --- & 1.35 & 2.17  & 295.3 & 0.79 & 0.65 \\
  MAR-B~\citep{mar}\diffm
    & M+D 208M      & 0.87 & --- & 2.31 & 3.48  & 281.7 & 0.82 & 0.57 \\
  MAR-H~\citep{mar}\diffm
    & M+D 943M      & 0.87 & --- & 1.55 & 2.35  & 303.7 & 0.81 & 0.62 \\
  FlowAR-B~\citep{flowar}\diffm
    & Flow+VAR 300M & 0.87 & --- & 2.90 & ---   & 272.5 & 0.84 & 0.54 \\

  \midrule
  \multicolumn{9}{l}{\textit{Discrete tokens, next-scale and masked generation}} \\
  VAR-d30~\citep{var}
    & VAR 2.0B      & ---  & --- & 1.92 & ---   & 323.1 & 0.82 & 0.59 \\
  MaskGIT~\citep{maskgit}
    & Mask.\ 227M   & 2.28 & --- & ---  & 6.18  & 182.1 & 0.80 & 0.51 \\
  MAGVIT-v2~\citep{magvitv2}
    & Mask.\ 307M   & ---  & --- & 1.78 & 3.65  & 319.4 & ---  & ---  \\

  \midrule
  \multicolumn{9}{l}{\textit{2D discrete tokens, AR}} \\
  VQGAN~\citep{vqgan}
    & AR 1.4B       & 4.98 & --- & ---  & 15.78 & 74.3  & ---  & ---  \\
  RQ-Trans.~\citep{rqtransformer}
    & AR 3.8B       & 3.20 & --- & ---  & 7.55  & 134.0 & ---  & ---  \\
  LlamaGen-XL~\citep{llamagen}
    & AR 775M       & 0.94 & --- & 2.62 & ---   & 244.1 & 0.80 & 0.57 \\
  LlamaGen-XXL~\citep{llamagen}
    & AR 1.4B       & 0.94 & --- & 2.34 & ---   & 253.9 & 0.80 & 0.59 \\
  PAR-L-4$\times$~\citep{par}
    & PAR 343M      & 0.94 & --- & 3.76 & ---   & 218.9 & 0.84 & 0.50 \\

  \midrule
  \multicolumn{9}{l}{\textit{1D discrete tokens, non-standard AR (mask / random-order / hierarchical / parallel)}} \\
  TiTok-S-128~\citep{titok}
    & Mask.\ 287M   & 1.71 & --- & 1.97 & 4.44 & 281.8 & ---  & ---  \\
  TiTok-L-32~\citep{titok}
    & Mask.\ 177M   & 2.21 & --- & 2.77 & 3.15 & 199.8 & ---  & ---  \\
  ImageFolder~\citep{imagefolder}\sem\drop
    & VAR 362M      & 0.80 & --- & 2.60 & ---  & 295.0 & 0.75 & 0.63 \\
  ResTok~\citep{restok}\sem\drop
    & HAR 326M      & 1.28 & --- & 2.34 & ---  & 257.8 & 0.79 & 0.60 \\
  SpectralAR~\citep{spectralar}
    & AR 310M       & 4.03 & --- & 3.02 & ---  & 282.2 & 0.81 & 0.55 \\
  DetailFlow-16~\citep{detailflow}
    & PAR 326M      & 1.22 & 19.05 & 2.96 & --- & 221.4 & 0.82 & 0.57 \\
  RAR-XL~\citep{rar}
    & RandAR 955M   & ---  & --- & 1.50 & 3.72 & 306.9 & 0.80 & 0.62 \\
  RAR-XXL~\citep{rar}
    & RandAR 1.5B   & ---  & --- & 1.48 & 3.26 & 326.0 & 0.80 & 0.63 \\

  \midrule
  \multicolumn{9}{l}{\textit{1D discrete tokens, standard AR}} \\
  GigaTok-B-L~\citep{gigatoken}\sem
    & AR 111M       & 0.81 & 21.21 & 3.26 & ---  & 221.0 & 0.81 & 0.56 \\
  VFMTok-L~\citep{vfmtok}\sem
    & AR 343M       & 0.89 & --- & 2.75 & 2.11 & 278.8 & 0.84 & 0.57 \\
  AliTok-B~\citep{alitok}\clus
    & AR 177M       & 0.86 & --- & 1.44 & 2.40 & 319.5 & 0.77 & 0.65 \\
  AliTok-L~\citep{alitok}\clus
    & AR 318M       & 0.86 & --- & 1.38 & 1.98 & 326.2 & 0.78 & 0.65 \\
  AliTok-XL~\citep{alitok}\clus
    & AR 662M       & 0.86 & --- & 1.28 & 1.88 & 306.3 & 0.79 & 0.65 \\

  \addlinespace
  \textbf{VCQ B-B  (ours)}
    & AR 115M       & 2.77 & 20.02 & 4.79 & 14.80 & 221.9 & 0.828 & 0.452 \\
  \textbf{VCQ B-L  (ours)}
    & AR 343M       & 2.77 & 20.02 & 2.75 & 7.47  & 258.3 & 0.801 & 0.572 \\
  \textbf{VCQ B-XL (ours)}
    & AR 684M       & 2.77 & 20.02 & 2.38 & 6.82  & 306.9 & 0.802 & 0.587 \\
  \textbf{VCQ L-B  (ours)}
    & AR 115M       & 0.96 & 19.71 & 2.73 & 5.14  & 218.7 & 0.777 & 0.590 \\
  \textbf{VCQ L-L  (ours)}
    & AR 343M       & 0.96 & 19.71 & 1.90 & 2.72  & 243.6 & 0.752 & 0.658 \\
  \textbf{VCQ L-XL (ours)}
    & AR 684M       & 0.96 & 19.71 & 1.71 & 2.29  & 250.8 & 0.762 & 0.670 \\
  \bottomrule
\end{tabular}
\end{center}
\end{table}

\subsection{Baseline comparison: schedule selection and ablations}
\label{sec:generation}

Table~\ref{tab:internal} compares all codebook configurations under the same Base Tokenizer + Base AR. The training objective, optimizer, and number of steps are identical; the only difference is the codebook structure.

\paragraph{Variable structure vs.\ total capacity.}
Constant $K$=8192 and VCQ-Linear have similar reconstruction quality (rFID 2.36 vs.\ 2.42), while gFID w/o CFG improves from 18.16 to 14.70. Their total nominal capacities, $C=\sum_t\log_2K_t$, are approximately 3328 and 3209 bits per image, respectively, a 3.6\% difference. BPP is $C/(256\times256)$. Capacity alone also does not order the observed results: Power has lower BPP than Cosine (0.041 vs.\ 0.044) but worse gFID w/o CFG (16.19 vs.\ 14.80). Together, these results support VCQ's combination of a strong early information bottleneck and larger codebooks at later positions.

\paragraph{Schedule selection.}
From Constant ($K$=16384) to VCQ-Cosine, gFID w/o CFG drops from 27.98 to 14.80 ($-$47\%) and gFID drops from 6.43 to 4.79 ($-$26\%). VCQ-Linear matches VCQ-Cosine on gFID w/o CFG (14.70 vs.\ 14.80), but Cosine still wins under inference-time guidance (4.79 vs.\ 5.32 gFID). We therefore use Cosine as the default for the full pipeline.

\paragraph{Reconstruction cost vs.\ generation gain.}
VCQ-Cosine improves generation at a reconstruction cost: rFID increases from 2.26 to 2.77, while gFID w/o CFG decreases from 27.98 to 14.80. Linear retains better reconstruction quality, while Cosine gives the best guided generation in our comparison.

\subsection{Comparison with existing methods}

Table~\ref{tab:gen} jointly compares VCQ against existing methods on reconstruction (rFID/PSNR) and generation (gFID/IS/Prec/Rec). The comparison groups span continuous-token diffusion baselines, VAR/masked-transformer baselines, 2D discrete-token AR, and 1D discrete-token AR. On the reconstruction side, VCQ-L reaches an rFID of 0.96, competitive among 256-token methods. On the generation side, VCQ uses no auxiliary training techniques (no semantic regularization~\citep{gigatoken,vfmtok}, no causal alignment~\citep{alitok}, no hierarchical residuals~\citep{restok}, no random-order training~\citep{rar,randar}). Our controlled comparison in Table~\ref{tab:internal} shows that VCQ improves gFID w/o CFG under the same AR training configuration.

\paragraph{Scaling behavior.}
VCQ's generation quality improves as the tokenizer and AR model scale up. With the Base Tokenizer, VCQ B-B (115M) already cuts gFID w/o CFG from 27.98 to 14.80 at matched scale; scaling the AR alone, VCQ B-L drops it to 7.47, and VCQ B-XL to 6.82. Using the Large-tokenizer configuration in Appendix~\ref{app:training}, VCQ L-B reaches a gFID w/o CFG of 5.14 (already better than any Base-Tok configuration); L-L and L-XL further drive gFID w/o CFG down to 2.72 and \textbf{2.29}, on par with strong CFG-equipped 2D-token AR baselines such as LlamaGen-XXL (1.4B, gFID 2.34). With CodebookSize-Aware CFG, VCQ L-L and VCQ L-XL reach gFID 1.90 and \textbf{1.71} with 343M and 684M AR parameters, respectively, and no auxiliary training techniques.

\begin{figure}[t]
\centering
\includegraphics[width=0.92\textwidth]{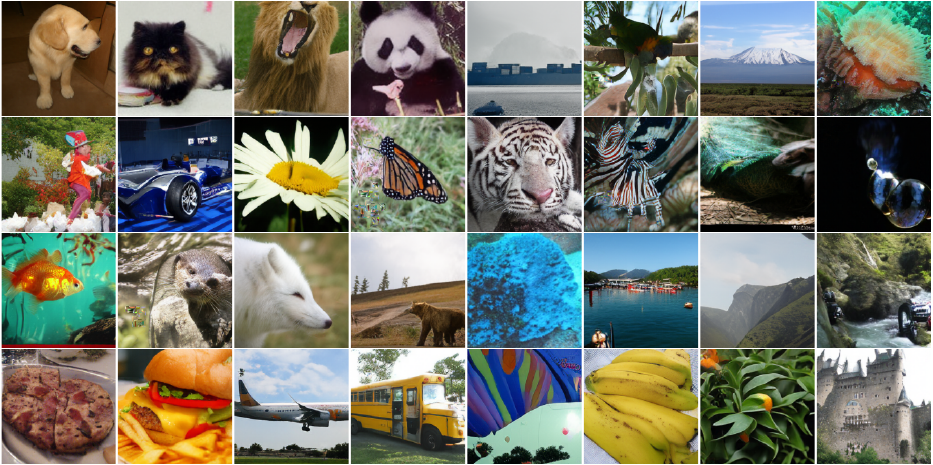}
\caption{\textbf{Uncurated class-conditional samples from VCQ L-XL.}
A $4{\times}8$ grid covering visually diverse ImageNet classes with no manual curation.
Samples use the Large Tokenizer ($K_{\max}\!=\!11264$) and AR-XL (684M) with CodebookSize-Aware CFG at the best gFID configuration: cosine CFG with $s=3$ and $p=1.0$.}
\label{fig:samples}
\end{figure}
\subsection{Semantic aggregation and visualization}
\label{sec:semantic}

We average the pre-quantization features of the first $k$ tokens and train a linear classifier on the resulting feature, following the linear probing protocol of MAE~\citep{mae}.

\begin{table}[t]
  \caption{\textbf{Semantic aggregation.} Linear probing uses mean-pooled pre-quantization features from the first $k$ positions.}
  \label{tab:semantic}
  \begin{center}
\small
\begin{tabular}{@{}lccc@{}}
    \toprule
    Tokenizer & $k$ & Top-1 $\uparrow$ & Top-5 $\uparrow$ \\
    \midrule
    Constant ($K$=16384) & 10  & 27.12\% & 49.27\% \\
    Constant ($K$=16384) & 256 & 48.02\% & 71.51\% \\
    \addlinespace
    VCQ-Linear   & 10  & 31.24\% & 54.59\% \\
    VCQ-Linear   & 256 & 46.60\% & 70.11\% \\
    \addlinespace
    VCQ-Cosine   & 10  & 43.77\% & 67.70\% \\
    VCQ-Cosine   & 256 & \textbf{48.79\%} & \textbf{72.70\%} \\
    \addlinespace
    VCQ-Power ($\alpha$=2.5) & 10  & \textbf{49.42\%} & \textbf{72.70\%} \\
    VCQ-Power ($\alpha$=2.5) & 256 & 46.93\% & 70.71\% \\
    \bottomrule
  \end{tabular}
\end{center}
\end{table}

$K_{\min}\!=\!2$ limits the first discrete code to a capacity of only 1 bit. The probing results suggest that this bottleneck encourages semantic information to concentrate in early encoder features. Using features from only 10 positions (3.9\% of the sequence), VCQ-Cosine achieves a linear probing accuracy of 43.77\%, while Constant reaches only 27.12\%. With all 256 positions, the linear-probe accuracies are comparable (48.79\% vs.\ 48.02\%), while VCQ performs better with only the first 10 positions.

The Cosine schedule grows more slowly at the start than Linear, imposing a stronger early information bottleneck; the first-10-position accuracy differs by 12.5 percentage points. This semantic concentration emerges without explicit semantic supervision.

\section{Conclusion}

We identify the Entropy Cliff: large uniform codebooks can consume the finite-data entropy budget within the first few positions, leaving most empirical continuations fixed. VCQ delays this decline by using small codebooks at early positions and progressively larger ones later. This simple structural change redistributes capacity along the sequence without changing the loss function, parameter count, or AR training procedure.

On ImageNet, VCQ reduces gFID w/o CFG from 27.98 to 14.80 under the same AR training configuration. Combined with CodebookSize-Aware CFG, it improves gFID from 6.43 to 4.79. Scaling to the Large tokenizer and a 684M-parameter AR model further brings gFID to 1.71, without semantic regularization or causal alignment. The strong information bottleneck at early positions also encourages semantic concentration in encoder features: linear probing on the first 10 positions reaches 43.8\% accuracy, compared with 27.1\% for the uniform codebook.

Our experiments focus on ImageNet $256\times256$ class-conditional generation, and the empirical cliff depends on the tokenized data and codebook schedule. Evaluating VCQ on more datasets and combining it with existing training techniques are natural next steps. These results show that codebook design should consider both its total capacity and how that capacity is distributed across the sequence.

\subsection*{AI use statement}
LLMs were used for language polishing, coding assistance, reviewing mathematical formulas, and searching for related literature. The core method and research were developed without AI assistance. The authors take full responsibility for the final content of this paper.

\bibliographystyle{unsrtnat}
\bibliography{references}
\clearpage
\appendix

\section{Related work}
\label{app:related}

\textbf{Discrete visual tokenizers.}
VQ-VAE~\citep{vqvae} and VQ-GAN~\citep{vqgan} established the basic paradigm of representing images as sequences of discrete codes from a finite codebook. Subsequent works have improved from various angles: better optimization and larger codebooks (ViT-VQGAN~\citep{vitvqgan}; LlamaGen~\citep{llamagen} with $K\!=\!16384$; Open-MAGVIT2~\citep{openmagvit2} and MAGVIT-v2~\citep{magvitv2} with LFQ; IBQ~\citep{ibq} with index back-propagation); 1D / variable-length tokenizers (TiTok~\citep{titok}; FlexTok~\citep{flextok}; One-D-Piece~\citep{onedpiece}; SelfTok~\citep{selftok}); training techniques (GigaTok~\citep{gigatoken} with semantic regularization, AliTok~\citep{alitok} with causal alignment, ResTok~\citep{restok} with hierarchical residuals, ImageFolder~\citep{imagefolder} with folded tokens, MaskBit~\citep{maskbit} with bit-level codes, FlowMo~\citep{flowmo} with mode-seeking diffusion autoencoders); and the use of pre-trained vision foundation models such as DINOv2~\citep{dinov2} and CLIP~\citep{clip} as encoders or alignment targets (LightningDiT/VA-VAE~\citep{lightningdit}, VFMTok~\citep{vfmtok}). To the best of our knowledge, VCQ is the first work to systematically explore position-dependent codebook cardinality $K_t$ as a primary design variable.

\textbf{AR visual generation.}
Early AR image models such as DALL-E~\citep{dalle} and Parti~\citep{parti} treat tokenized images as sequences and generate them with large Transformers~\citep{transformer}; VQ-GAN~\citep{vqgan} brought GPT-style class-conditional AR to ImageNet. Subsequent developments include scaling (LlamaGen~\citep{llamagen}), residual quantization (RQ-Transformer~\citep{rqtransformer}), next-scale prediction (VAR~\citep{var}), masked / non-causal generation (MaskGIT~\citep{maskgit}), diffusion-loss heads on continuous tokens (MAR~\citep{mar}, FlowAR~\citep{flowar}), parallel decoding (PAR~\citep{par}), random or randomized generation orders (RandAR~\citep{randar}, RAR~\citep{rar}), and coarse-to-fine 1D ordering (DetailFlow~\citep{detailflow}, SpectralAR~\citep{spectralar}). On the diffusion side, ADM~\citep{adm} and DDPM~\citep{ddpm} together with classifier-free guidance~\citep{cfg} laid the foundation for latent diffusion (LDM~\citep{ldm}) and Diffusion/Interpolant Transformers (DiT~\citep{dit}, SiT~\citep{sit}, REPA~\citep{repa}, LightningDiT~\citep{lightningdit}). We study codebook structure with 1D token sequences and standard next-token prediction.

\textbf{Reconstruction-generation trade-off.}
GigaTok~\citep{gigatoken} attributes the harm of increasing $K$ for generation to increased latent space complexity, and proposes semantic regularization as a remedy. We provide a complementary information-theoretic explanation: uniform large codebooks can leave almost no empirical conditional entropy at later positions, while regularization allows the AR model to learn a smoother distribution. VCQ changes this data-side allocation of information.

\textbf{Evaluation protocol.}
Throughout the paper we follow standard protocols: FID~\citep{fid} and Inception Score~\citep{is} as fidelity/quality summaries, and the improved Precision/Recall~\citep{precrecall} as a separate measure of fidelity vs.\ coverage. Reconstruction quality is reported with rFID, PSNR. We use the standard ADM~\citep{adm} evaluation implementation for FID, IS, and Precision/Recall. Architectural and optimization choices follow widely-used recipes (ViT~\citep{vit}, Transformer~\citep{transformer}, AdamW~\citep{adamw}, PatchGAN-style adversarial loss~\citep{pix2pix}, Llama-style decoder-only AR~\citep{llama}).

\section{Training details}
\label{app:training}
\paragraph{Dataset and compute accounting.}
All tokenizer and autoregressive (AR) experiments are trained on the ImageNet-1K training set.
Experiments were run on a heterogeneous NVIDIA GPU cluster.
For comparability, we report training duration as A100-equivalent wall-clock time.

\paragraph{Tokenizer training.}
Both tokenizers use a 1D sequence length of 256 and a shared cosine VCQ codebook with $K_{\min}=2$.
The Base tokenizer uses $K_{\max}=16384$, while the Large tokenizer uses $K_{\max}=11264$.
The Base tokenizer follows a ViT-B encoder and ViT-B decoder design with 172M parameters in total.
The Large tokenizer keeps the same ViT-B encoder but uses a larger ViT-L decoder, resulting in 496M parameters.
For the Large tokenizer, LPIPS is computed using a ConvNeXt-S backbone following~\citep{alitok}.

Both tokenizers are trained for 130 ImageNet epochs with a global batch size of 256.
The 1.3M optimization-step count includes alternating generator and discriminator updates, corresponding to approximately 650k tokenizer (generator) updates.
The Base tokenizer is trained on 8 GPUs and takes 13.0 A100-equivalent hours.
The Large tokenizer is trained on 8 GPUs and takes 18.2 A100-equivalent hours.

Tokenizer optimization uses AdamW.
After a 13k-step warmup, the learning rate stays at $10^{-4}$ until step 65k, then decays linearly to $3\times10^{-5}$.
The reconstruction objective combines L1 reconstruction, LPIPS perceptual loss, adversarial generator loss, adversarial discriminator loss, and the VQ commitment loss.
The L1, LPIPS, discriminator, and commitment losses use weight 1.0, while the adversarial generator loss uses weight 0.1.
We use random cropping and horizontal flipping for image augmentation, and adversarial augmentation is always enabled.
Training uses bf16 mixed precision.

\paragraph{Training budgets across methods.}
Table~\ref{tab:training_budgets} summarizes the training epochs for VCQ, GigaTok~\citep{gigatoken}, and AliTok~\citep{alitok}. VCQ uses 130 tokenizer epochs, compared with 200 for GigaTok and 180 for AliTok. The AR budgets depend on the configuration; our Base-tokenizer and Large-tokenizer groups use 400 and 800 epochs, respectively. Model sizes, batch sizes, and training recipes vary across methods.

\begin{table}[ht]
\caption{Training budgets in epochs. VCQ tokenizer epochs count generator updates. AliTok AR budgets are listed for B/L/XL; VCQ AR budgets apply to all three AR sizes within each tokenizer group.}
\label{tab:training_budgets}
\begin{center}
\small
\begin{tabular}{@{}lcc@{}}
\toprule
Method & Tokenizer epochs & AR epochs \\
\midrule
GigaTok & 200 & 300 \\
AliTok & 180 & 800 / 800 / 400 \\
VCQ Base tokenizer & 130 & 400 \\
VCQ Large tokenizer & 130 & 800 \\
\bottomrule
\end{tabular}
\end{center}
\end{table}

\paragraph{Autoregressive training.}
AR models are trained on frozen tokenizers with standard next-token prediction.
All AR runs use ten-crop augmentation, horizontal flipping, class-label dropout of 0.1, AR dropout of 0.1, and token-embedding dropout of 0.1.

For the Base-tokenizer group, BB, BL, and BXL are trained for 1.0M optimization steps with a global batch size of 512, corresponding to 400 ImageNet epochs.
BB uses the 115M AR-B model with 12 layers, 12 heads, and hidden dimension 768.
It is trained on 8 GPUs and takes 7.6 A100-equivalent hours.
BL uses the 343M AR-L model with 24 layers, 16 heads, and hidden dimension 1024.
It is trained on 8 GPUs and takes 15.7 A100-equivalent hours.
BXL uses the 684M AR-XL model with 32 layers, 20 heads, and hidden dimension 1280.
It is trained on 16 GPUs and takes 20.0 A100-equivalent hours.

For this Base-tokenizer group, AR optimization uses AdamW with $\beta=(0.9,0.96)$.
After a 10k-step warmup, the learning rate stays at $2\times10^{-4}$ until step 50k, then decays linearly to $4\times10^{-5}$.
The AR weight decay is $3\times10^{-2}$, while bias, normalization, and adaptive normalization parameters are excluded from weight decay.
Gradient clipping is set to 1.0.

For the Large-tokenizer group, LB, LL, and LXL are trained for 500k optimization steps with a global batch size of 2048, corresponding to 800 ImageNet epochs.
LB uses the 115M AR-B model and is trained on 16 GPUs for 5.3 A100-equivalent hours.
LL uses the 343M AR-L model and is trained on 16 GPUs for 12.4 A100-equivalent hours.
LXL uses the 684M AR-XL model and is trained on 32 GPUs for 15.1 A100-equivalent hours.

For this Large-tokenizer group, AR optimization also uses AdamW with $\beta=(0.9,0.96)$ and the same weight-decay rule as above.
The learning rate is linearly decayed from $4\times10^{-4}$ to $10^{-5}$, with a 62.5k-step warmup.
Gradient clipping is set to 1.0.

\section{Inference details}
\label{app:inference}

Table~\ref{tab:inference} lists the CFG and temperature settings used for the main gFID and gFID w/o CFG results.

\paragraph{Cosine-power CFG.}
We use the power-cosine profile to schedule the base guidance scale:
\begin{equation}
  c_p(\tau)=\frac{1-\cos(\pi\tau^p)}{2},\qquad
  \bar{s}_t=a+(s-a)c_p(\tau_t),
  \label{eq:cfg_cosine_power}
\end{equation}
where $\tau_t\in[0,1]$ is normalized generation progress. The parameters $s$ and $p>0$ correspond to scale and p in Table~\ref{tab:inference}; larger $p$ delays the rise in guidance. Constant codebooks use $s_t=\bar{s}_t$, while VCQ applies the codebook weight in Eq.~\eqref{eq:bitcfg}. All scales follow the logits convention $(1+s_t)\,\text{logits}_{\text{cond}}-s_t\,\text{logits}_{\text{uncond}}$. Runs without CFG set $s_t=0$ throughout.

\begin{table}[t]
\caption{CFG and temperature configurations corresponding to the reported gFID and gFID w/o CFG values.}
\label{tab:inference}
\begin{center}
\small
\begin{tabular}{@{}llll@{}}
\toprule
Model & Comment & CFG & Temperature \\
\midrule
Constant ($K$=16384) & gFID=6.43 & cosine, scale=20, p=1.5 & 1.0 \\
Constant ($K$=16384) & gFID w/o CFG=27.98 & cfg=0 & 0.85 \\
\addlinespace
Constant ($K$=8192) & gFID=5.87 & cosine, scale=20, p=2.0 & 1.0 \\
Constant ($K$=8192) & gFID w/o CFG=18.16 & cfg=0 & 0.85 \\
\addlinespace
VCQ-Linear & gFID=5.32 & cosine, scale=14, p=1.75 & 1.0 \\
VCQ-Linear & gFID w/o CFG=14.70 & cfg=0 & 0.85 \\
\addlinespace
VCQ-Cosine / VCQ B-B & gFID=4.79 & cosine, scale=10, p=1.5 & 1.0 \\
VCQ-Cosine / VCQ B-B & gFID w/o CFG=14.80 & cfg=0 & 0.85 \\
\addlinespace
VCQ-Power ($\alpha$=2.5) & gFID=5.14 & cosine, scale=10, p=1.75 & 1.0 \\
VCQ-Power ($\alpha$=2.5) & gFID w/o CFG=16.19 & cfg=0 & 0.85 \\
\addlinespace
VCQ B-L & gFID=2.75 & cosine, scale=4, p=0.75 & 1.0 \\
VCQ B-L & gFID w/o CFG=7.47 & cfg=0 & 0.85 \\
\addlinespace
VCQ B-XL & gFID=2.38 & cosine, scale=5, p=0.75 & 1.0 \\
VCQ B-XL & gFID w/o CFG=6.82 & cfg=0 & 0.85 \\
\addlinespace
VCQ L-B & gFID=2.73 & cosine, scale=4, p=1.25 & 1.0 \\
VCQ L-B & gFID w/o CFG=5.14 & cfg=0 & 0.90 \\
\addlinespace
VCQ L-L & gFID=1.90 & cosine, scale=4, p=1.25 & 1.0 \\
VCQ L-L & gFID w/o CFG=2.72 & cfg=0 & 0.90 \\
\addlinespace
VCQ L-XL & gFID=1.71 & cosine, scale=3, p=1.0 & 1.0 \\
VCQ L-XL & gFID w/o CFG=2.29 & cfg=0 & 0.95 \\
\bottomrule
\end{tabular}
\end{center}

\end{table}

\section{Disentangling CodebookSize-Aware CFG and VCQ contributions}
\label{app:posaware}

The gFID improvement of VCQ involves two factors: the codebook structure (VCQ) and the inference strategy (CodebookSize-Aware CFG from Eq.~\eqref{eq:bitcfg}). This section disentangles them through two controlled experiments.

\textbf{Experiment 1: Vanilla CFG vs.\ CodebookSize-Aware CFG.}
Table~\ref{tab:disentangle} switches inference strategies on the same VCQ tokenizer and AR model. With other sampling settings fixed, gFID w/o CFG is unaffected by the choice of CFG schedule and reflects the AR model's modeling quality.

\begin{table}[h]
  \caption{\textbf{Disentangling VCQ and CodebookSize-Aware CFG contributions.} The VCQ rows use the same tokenizer and AR model with different inference strategies.}
  \label{tab:disentangle}
  \begin{center}
\small
\begin{tabular}{llccc}
    \toprule
    Tokenizer & Inference strategy & gFID $\downarrow$ & gFID\textsubscript{noCFG} $\downarrow$ & $\Delta$gFID \\
    \midrule
    Constant ($K$=16384)  & Cosine CFG  & 6.43  & 27.98 & --- \\
    Constant ($K$=8192)   & Cosine CFG ($a=0$) & 5.87  & 18.16 & --- \\
    \addlinespace
    VCQ                   & Cosine CFG & 7.61  & 14.80 & --- \\
    VCQ                   & CodebookSize-Aware CFG   & \textbf{4.79}  & 14.80 & $-$2.82 \\
    \bottomrule
  \end{tabular}
\end{center}
\end{table}

VCQ brings gFID w/o CFG from 27.98 to 14.80 ($-$47\%), which is the contribution of the codebook structure. CodebookSize-Aware CFG further reduces gFID from 7.61 to 4.79 ($-$2.82), which is the contribution of the inference strategy.

The gFID of VCQ plus Vanilla Cosine CFG (7.61) is worse than Constant (6.43) because Vanilla Cosine CFG does not match VCQ's variable structure: at early positions where the codebook is small and the distribution is concentrated, the standard cosine CFG schedule over-conditions. CodebookSize-Aware CFG ties the scale to excess information capacity (Eq.~\eqref{eq:bitcfg}); when $K_t = K_{\min}$, $s_t = 0$, resolving this mismatch.

\textbf{Experiment 2: Lower initial CFG on uniform codebooks.}
With a constant codebook, codebook size provides no position-dependent signal for guidance. We therefore compare standard cosine CFG, which starts at 1, with a schedule starting at 0. This allows CFG scales below 1 at early positions, as in CodebookSize-Aware CFG.

\begin{table}[h]
  \caption{\textbf{Lower initial CFG on uniform codebooks.}}
  \label{tab:posaware}
  \begin{center}
\small
\begin{tabular}{llcc}
    \toprule
    Tokenizer & CFG strategy & Best gFID $\downarrow$ & $\Delta$ \\
    \midrule
    Constant $K$=16K & Standard cosine CFG     & 6.43 & --- \\
    Constant $K$=16K & CodebookSize-Aware CFG      & 6.43 & 0.00 \\
    \addlinespace
    Constant $K$=8K  & Standard cosine CFG     & 5.92 & --- \\
    Constant $K$=8K  & CodebookSize-Aware CFG    & 5.87 & $-$0.05 \\
    \bottomrule
  \end{tabular}
\end{center}
\end{table}

Both CFG strategies give similar results on uniform codebooks, with no clear benefit from the lower initial scale. VCQ improves unguided generation, while CodebookSize-Aware CFG improves guided sampling with variable codebooks.

\section{Finite-data entropy analysis}
\label{app:entropy_theory}

\paragraph{Empirical distribution.}
Let $S$ be a uniformly sampled index among $N$ observations. Each observation is deterministically encoded as $X=(X_0,\ldots,X_{L-1})$, where $X_i$ takes at most $K_i$ values. All entropies in this section refer to this empirical distribution and use base-2 logarithms. Duplicate complete token sequences are allowed.

\paragraph{Remaining budget and proof of Proposition~\ref{prop:cliff}.}
Since the prefix $X_{<t}$ is a deterministic function of $S$, the chain rule gives
\begin{align}
 R(t)&=H(S\mid X_{<t})=\log_2N-H(X_{<t}) \\
 &=\log_2N-\sum_{i<t}H(X_i\mid X_{<i}).
\end{align}
Each conditional entropy satisfies $0\leq H(X_i\mid X_{<i})\leq\log_2K_i$, while $H(X_{<t})\leq H(S)=\log_2N$. Hence
\begin{equation}
 \max\left(0,\log_2N-\sum_{i<t}\log_2K_i\right)\leq R(t)\leq\log_2N.
\end{equation}
The remaining token sequence is also determined by $S$, so
\begin{align}
 H(X_{t:L}\mid X_{<t})
 &=H(X)-H(X_{<t}) \\
 &=R(t)-H(S\mid X)\leq R(t),
\end{align}
where $X_{t:L}$ denotes $(X_t,\ldots,X_{L-1})$. Thus $R(t)$ equals the remaining token entropy when all complete token sequences are distinct; otherwise it is an upper bound. Also, $R(t)=0$ if and only if all $N$ length-$t$ prefixes are distinct. In that case the entire tail is deterministic given the prefix. More generally, $R(t)\leq\epsilon$ leaves at most $\epsilon$ bits for the whole tail.

\paragraph{Conditional utilization.}
Define
\begin{equation}
 A_i=\min\bigl(\log_2K_i,R(i)\bigr),\qquad
 \rho_i=\begin{cases}
 H(X_i\mid X_{<i})/A_i,&A_i>0,\\
 0,&A_i=0.
 \end{cases}
\end{equation}
The tail bound above implies $0\leq H(X_i\mid X_{<i})\leq A_i$, so $\rho_i\in[0,1]$ and
\begin{equation}
 R(i+1)=R(i)-\rho_i A_i.
\end{equation}
If $\rho_i=1$ at each step with positive available budget, then
\begin{equation}
 R(t)=\max\left(0,\log_2N-\sum_{i<t}\log_2K_i\right).
\end{equation}
This expression gives the full-utilization floor.

\paragraph{Constant codebooks and the capacity floor.}
For $K_i=K$, the floor reaches zero at $t^*_{\mathrm{floor}}=\lceil\log_2N/\log_2K\rceil$. Equivalently, $K^t\geq N$ is necessary for the prefix space to distinguish all $N$ observations. It is not sufficient for the actual tokenizer to assign distinct prefixes. Under full conditional utilization the per-position entropy is
\begin{equation}
 H(X_t\mid X_{<t})=
 \min\left(\log_2K,\max\bigl(0,\log_2N-t\log_2K\bigr)\right).
\end{equation}
Without that condition, earlier positions may consume less entropy and leave more for later positions. For $N=K^2$ sequences of the form $(a,a,b)$, with $a$ and $b$ independently ranging over all $K$ codes, every position uses every code. Yet $H(X_1\mid X_0)=0$ and $H(X_2\mid X_0,X_1)=\log_2K$. Marginal usage is therefore insufficient to determine the cliff.

\paragraph{Dataset size.}
Table~\ref{tab:tstar} reports capacity floors computed from dataset size with $K=16384$.
\begin{table}[h]
\caption{Capacity floors for constant $K=16384$. Actual empirical curves depend on the tokenized samples and conditional utilization.}
\label{tab:tstar}
\begin{center}
\small
\begin{tabular}{@{}lrrc@{}}
\toprule
Dataset & $N$ & $\log_2N$ & $t^*_{\mathrm{floor}}$ \\
\midrule
CIFAR-10/100 & 50K & 15.6 & 2 \\
COCO & 118K & 16.9 & 2 \\
ImageNet-1K & 1.28M & 20.3 & 2 \\
CC12M & 12M & 23.5 & 2 \\
LAION-400M & 400M & 28.6 & 3 \\
LAION-5B & 5B & 32.2 & 3 \\
\bottomrule
\end{tabular}
\end{center}
\end{table}
Multiplying $N$ by $K$ increases this computed floor by one position.

\paragraph{Empirical estimation and model regularization.}
Equation~\eqref{eq:entropy_estimator} measures conditional entropy from observed prefix counts. For Figure~\ref{fig:entropy}(a), image prefixes are counted on the ImageNet validation set. A unique prefix has a single empirical continuation and contributes zero conditional entropy. The remaining-budget curve follows from these empirical entropies, whereas the capacity floor uses $N$ and the codebook schedule.

A regularized AR model can share information across prefixes and retain high predictive entropy. In our baseline, validation predictive entropy averages 13.5 bits over positions 3--16, and its minimum over positions is 9.1 bits. VCQ changes the empirical prefix structure, while regularization shapes the model distribution.

\paragraph{Training-set prefix counts.}
\label{app:prefix_counts}
We also compute prefix counts on all 1,281,167 ImageNet training images encoded with the constant codebook of $K=16384$. Table~\ref{tab:prefix_counts} reports the number of distinct length-$t$ prefixes and the number of images whose prefix is shared with another image.

\begin{table}[ht]
\caption{Prefix statistics for the constant codebook on the ImageNet training set. $t$ is the prefix length; mean count is the number of images per distinct prefix. The last column is the empirical entropy of the next token, in bits.}
\label{tab:prefix_counts}
\begin{center}
\small
\begin{tabular}{@{}crrrr@{}}
\toprule
$t$ & \shortstack{Distinct\\prefixes} & \shortstack{Images with\\shared prefixes} & \shortstack{Mean\\count} & $H_{\mathrm{emp}}(x_t\mid x_{<t})$ \\
\midrule
1 & 16,384 & 1,281,167 & 78.20 & 6.330 \\
2 & 1,251,312 & 42,963 & 1.024 & 0.0995 \\
3 & 1,274,113 & 13,495 & 1.006 & 0.00175 \\
4 & 1,274,808 & 12,463 & 1.005 & 0.00045 \\
\bottomrule
\end{tabular}
\end{center}
\end{table}

By prefix length 2, 96.6\% of images have unique prefixes; by length 3, this fraction reaches 98.9\%. The empirical next-token entropy falls below 0.002 bits at length 3. Thus most training prefixes have only one observed continuation, consistent with the data-side entropy cliff.

\section{Variable codebooks with a spatial 2D tokenizer}
\label{app:spatial_2d}
We also evaluate reconstruction with a spatial 2D tokenizer under constant and Cosine codebook schedules. Both configurations use the same training budget within this 2D setting.

\begin{table}[h]
\caption{Reconstruction with a spatial 2D tokenizer.}
\label{tab:spatial_2d}
\begin{center}
\small
\begin{tabular}{@{}lc@{}}
\toprule
Codebook schedule & rFID $\downarrow$ \\
\midrule
Constant & 2.16 \\
VCQ-Cosine & 3.92 \\
\bottomrule
\end{tabular}
\end{center}
\end{table}

The rFID increase is 1.76, compared with 0.51 for the main 1D comparison (2.26 to 2.77). In a spatial tokenizer, restricting early positions reduces the capacity available to particular spatial regions. The larger reconstruction cost suggests that the same schedule may be less suitable for this setting. Each schedule is compared with the constant-codebook baseline for the same tokenizer architecture.

\section{Additional generation samples}
\label{app:samples}
We provide the full uncurated sample gallery for \textbf{VCQ L-XL} (Large Tokenizer with $K_{\max}\!=\!11264$, AR-XL with 684M parameters).
Sampling uses CodebookSize-Aware CFG at the configuration that minimizes gFID under ADM protocol: cosine CFG with $s=3$ and $p=1.0$.
Each page is an $8{\times}8$ grid of 64 randomly selected ImageNet classes, one uncurated sample per class.
Across the eight appendix pages, the class sets are disjoint, yielding 512 sampled classes in total.
\begin{figure}[p]
\centering
\includegraphics[width=0.92\textwidth]{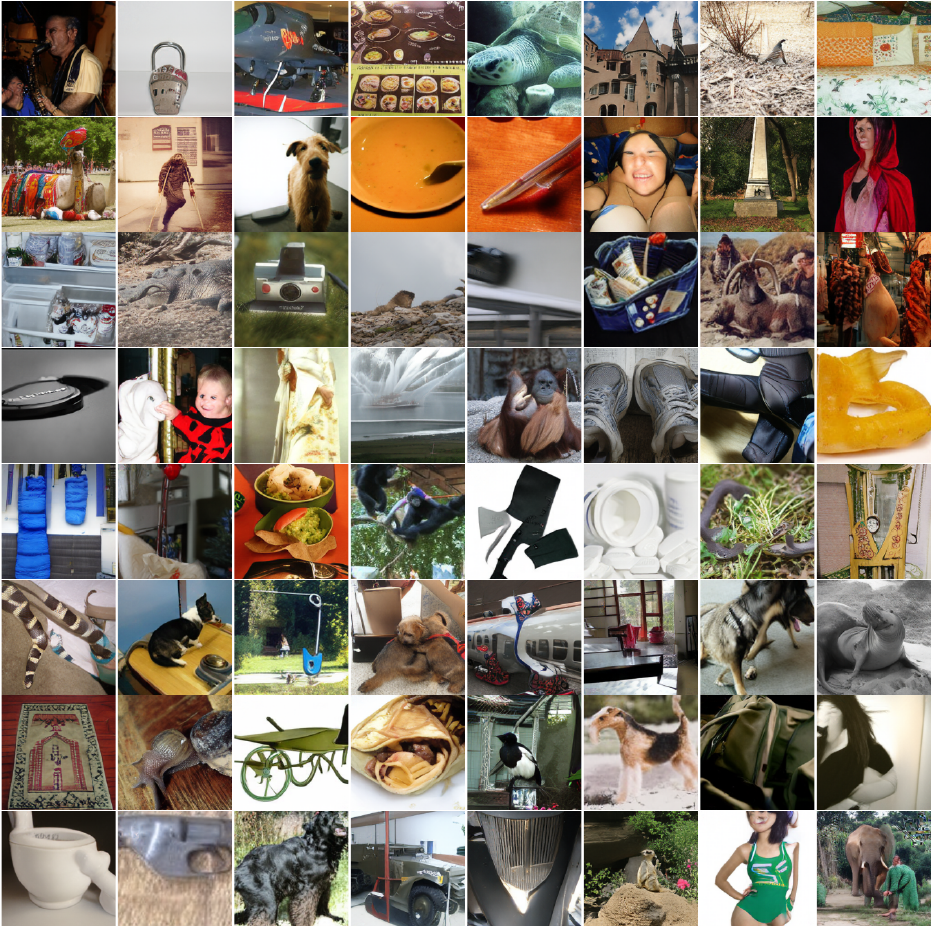}
\caption{\textbf{Uncurated VCQ L-XL samples (1/8).}
An $8{\times}8$ grid of 64 randomly selected distinct ImageNet classes. Sampling uses LXL with CodebookSize-Aware CFG, cosine CFG with $s=3$ and $p=1.0$.}
\label{fig:samples_app1}
\end{figure}
\begin{figure}[p]
\centering
\includegraphics[width=0.92\textwidth]{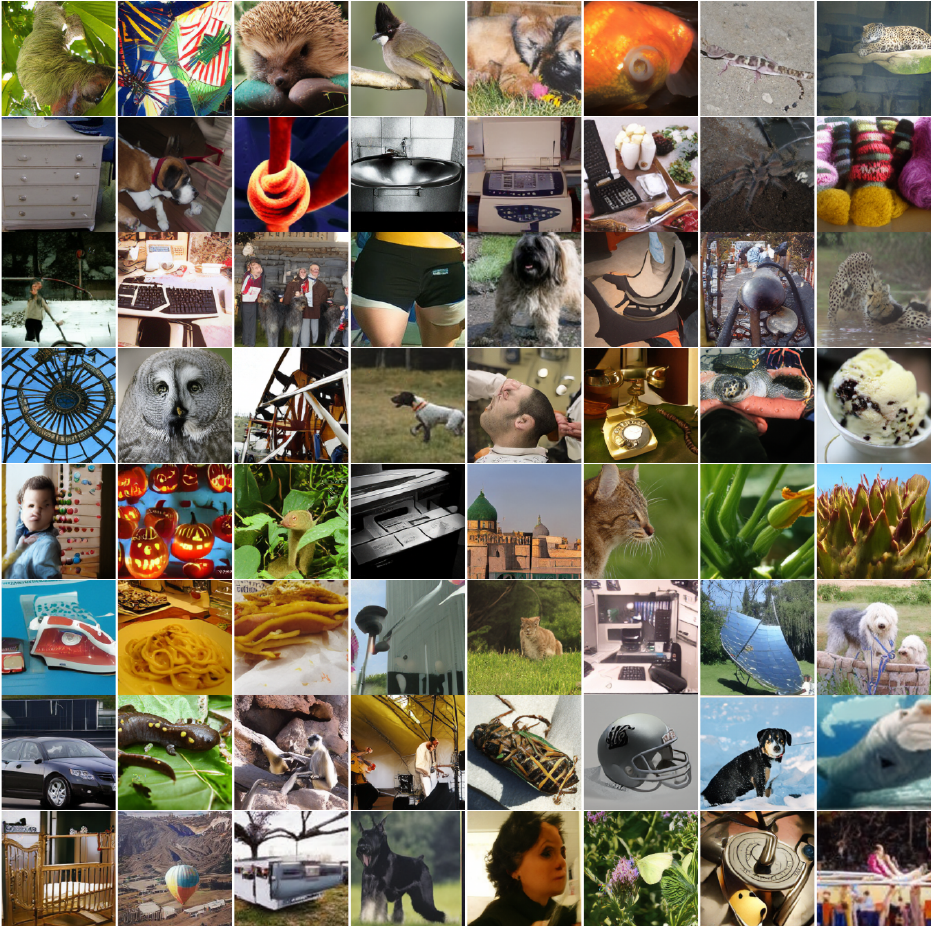}
\caption{\textbf{Uncurated VCQ L-XL samples (2/8).}
An $8{\times}8$ grid of 64 randomly selected distinct ImageNet classes, disjoint from the other gallery pages. Sampling uses LXL with CodebookSize-Aware CFG, cosine CFG with $s=3$ and $p=1.0$.}
\label{fig:samples_app2}
\end{figure}
\begin{figure}[p]
\centering
\includegraphics[width=0.92\textwidth]{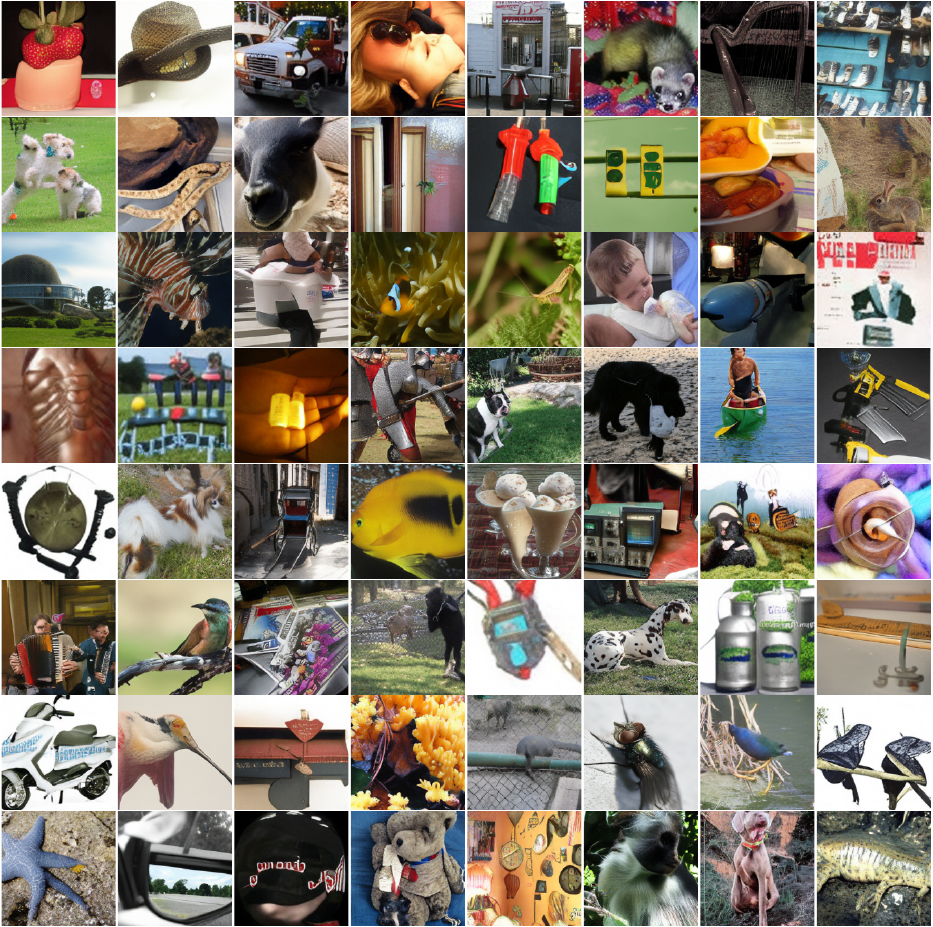}
\caption{\textbf{Uncurated VCQ L-XL samples (3/8).}
An $8{\times}8$ grid of 64 randomly selected distinct ImageNet classes, disjoint from the other gallery pages. Sampling uses LXL with CodebookSize-Aware CFG, cosine CFG with $s=3$ and $p=1.0$.}
\label{fig:samples_app3}
\end{figure}
\begin{figure}[p]
\centering
\includegraphics[width=0.92\textwidth]{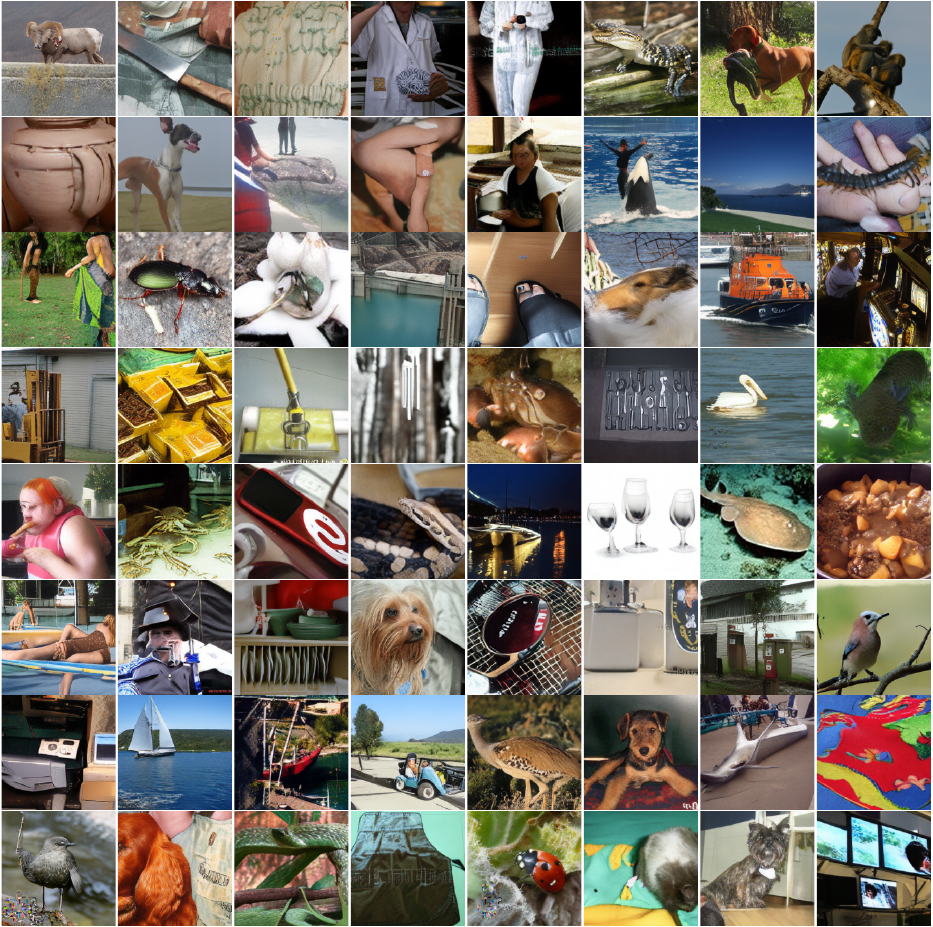}
\caption{\textbf{Uncurated VCQ L-XL samples (4/8).}
An $8{\times}8$ grid of 64 randomly selected distinct ImageNet classes, disjoint from the other gallery pages. Sampling uses LXL with CodebookSize-Aware CFG, cosine CFG with $s=3$ and $p=1.0$.}
\label{fig:samples_app4}
\end{figure}
\begin{figure}[p]
\centering
\includegraphics[width=0.92\textwidth]{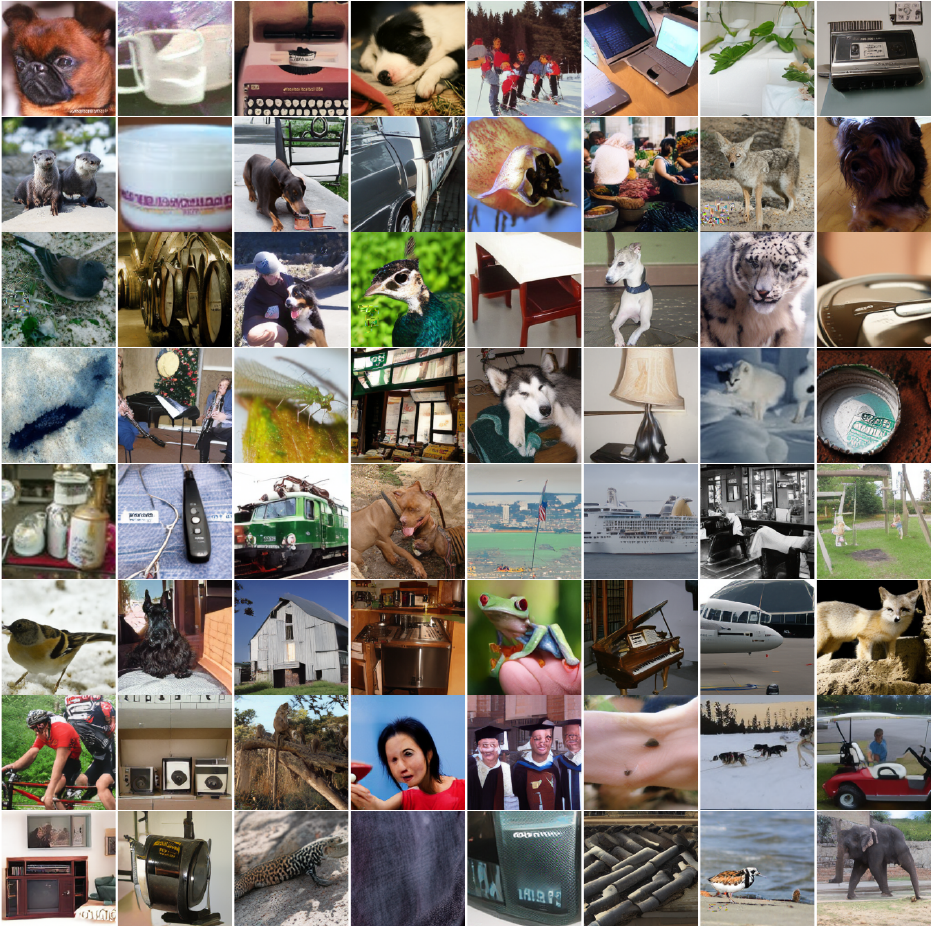}
\caption{\textbf{Uncurated VCQ L-XL samples (5/8).}
An $8{\times}8$ grid of 64 randomly selected distinct ImageNet classes, disjoint from the other gallery pages. Sampling uses LXL with CodebookSize-Aware CFG, cosine CFG with $s=3$ and $p=1.0$.}
\label{fig:samples_app5}
\end{figure}
\begin{figure}[p]
\centering
\includegraphics[width=0.92\textwidth]{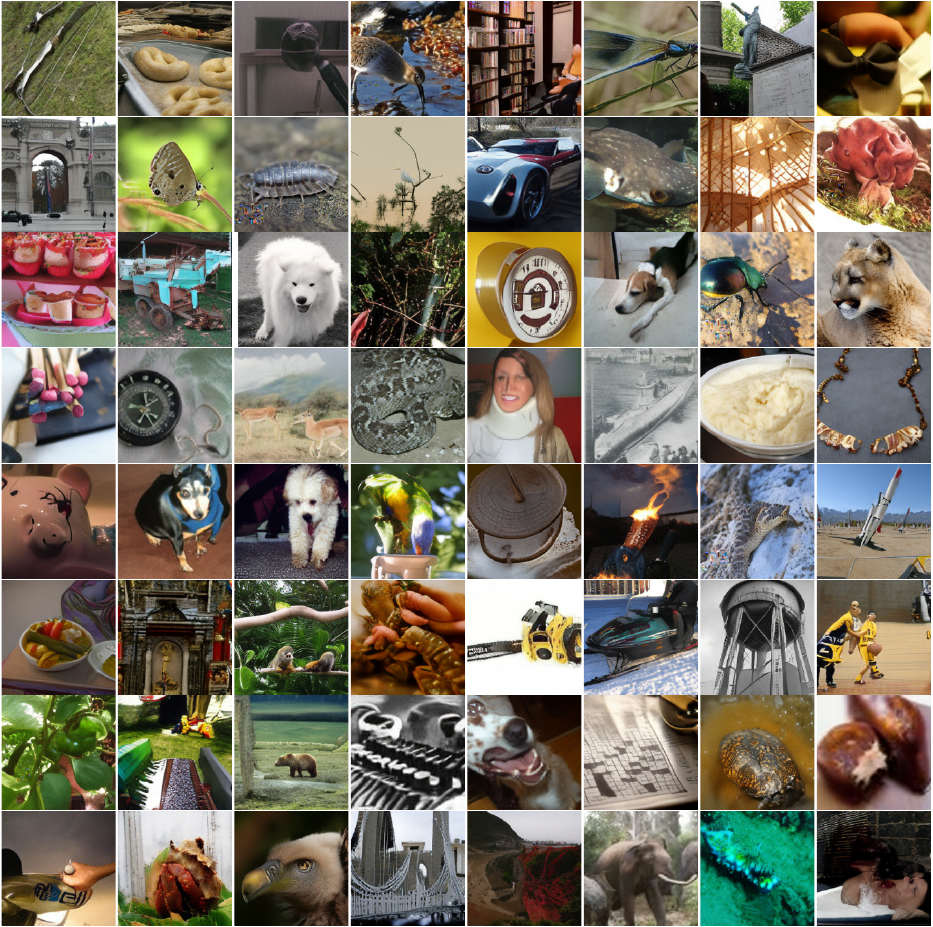}
\caption{\textbf{Uncurated VCQ L-XL samples (6/8).}
An $8{\times}8$ grid of 64 randomly selected distinct ImageNet classes, disjoint from the other gallery pages. Sampling uses LXL with CodebookSize-Aware CFG, cosine CFG with $s=3$ and $p=1.0$.}
\label{fig:samples_app6}
\end{figure}
\begin{figure}[p]
\centering
\includegraphics[width=0.92\textwidth]{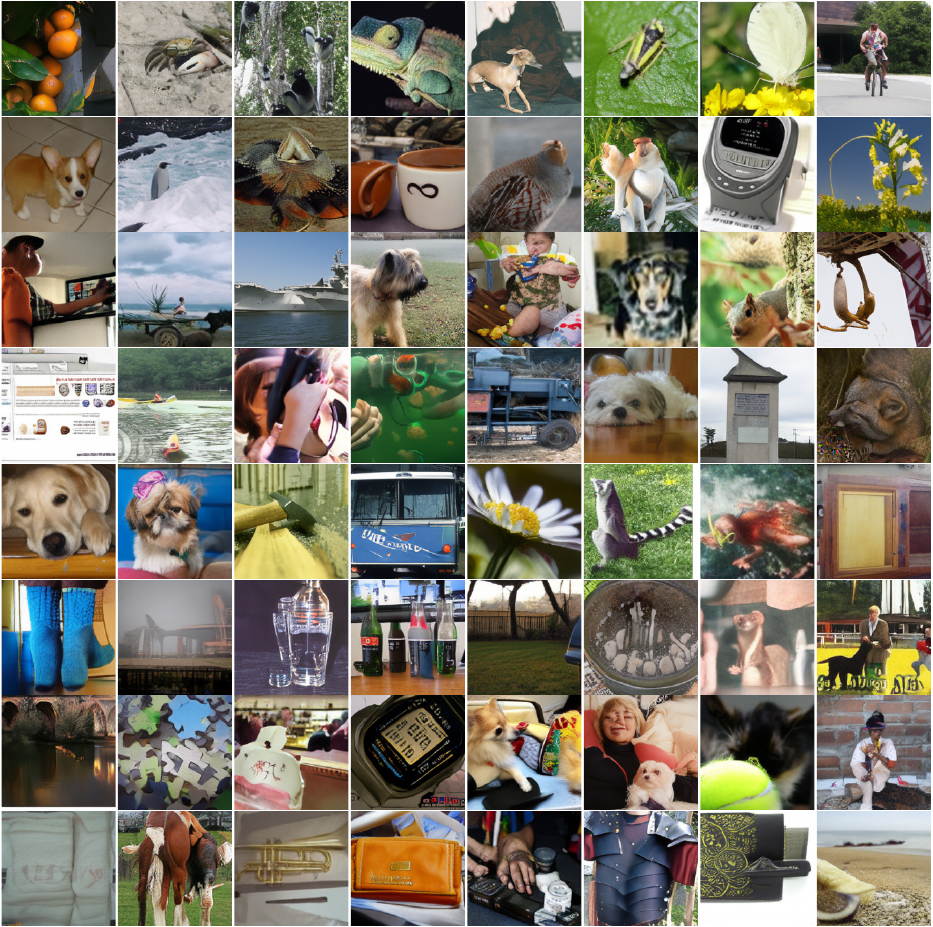}
\caption{\textbf{Uncurated VCQ L-XL samples (7/8).}
An $8{\times}8$ grid of 64 randomly selected distinct ImageNet classes, disjoint from the other gallery pages. Sampling uses LXL with CodebookSize-Aware CFG, cosine CFG with $s=3$ and $p=1.0$.}
\label{fig:samples_app7}
\end{figure}
\begin{figure}[p]
\centering
\includegraphics[width=0.92\textwidth]{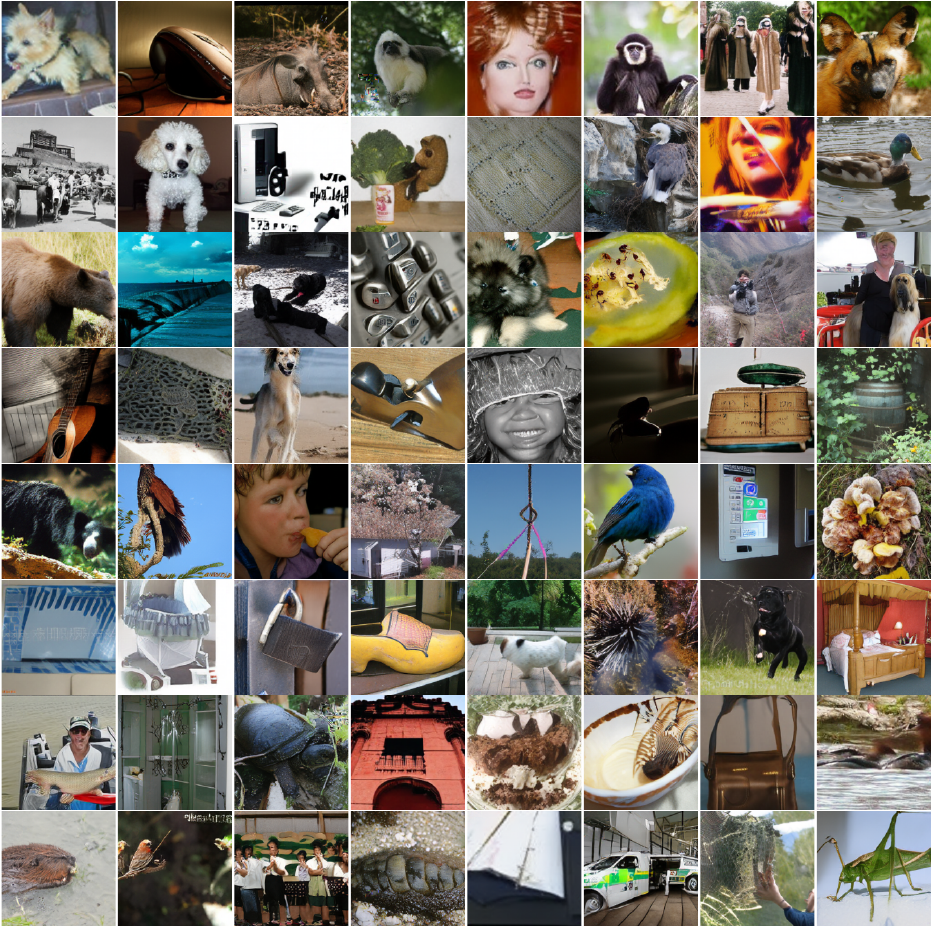}
\caption{\textbf{Uncurated VCQ L-XL samples (8/8).}
An $8{\times}8$ grid of 64 randomly selected distinct ImageNet classes, disjoint from the other gallery pages. Sampling uses LXL with CodebookSize-Aware CFG, cosine CFG with $s=3$ and $p=1.0$.}
\label{fig:samples_app8}
\end{figure}

\end{document}